\documentclass{article}
\usepackage{graphicx} 
\usepackage{tabularx}
\usepackage{amsmath}
\usepackage{xspace}
\usepackage{booktabs}
\usepackage{tcolorbox}
\usepackage{multirow}

\usepackage{colortbl}
\usepackage{graphicx}
\usepackage{enumitem}

\usepackage{wrapfig}

\usepackage{amsmath}
\usepackage{amsfonts}
\usepackage{bm}
\usepackage{vcell}

\usepackage[utf8]{inputenc} 
\usepackage[T1]{fontenc}    
\usepackage{hyperref}       
\usepackage{url}            
\usepackage{booktabs}       
\usepackage{amsfonts}       
\usepackage{nicefrac}       
\usepackage{microtype}      
\usepackage{xcolor}         

\usepackage{booktabs}
\usepackage{graphicx}
\usepackage{subcaption}
\usepackage{tabularx}
\usepackage{graphicx}
\usepackage{amsmath}
\usepackage{amssymb}
\usepackage[preprint]{corl_2024} 
\newcommand{\method}{\textsc{Aha-13B}\xspace}
\newcommand{\name}{\textsc{Aha}\xspace}

\title{\includegraphics[height=2.6ex]{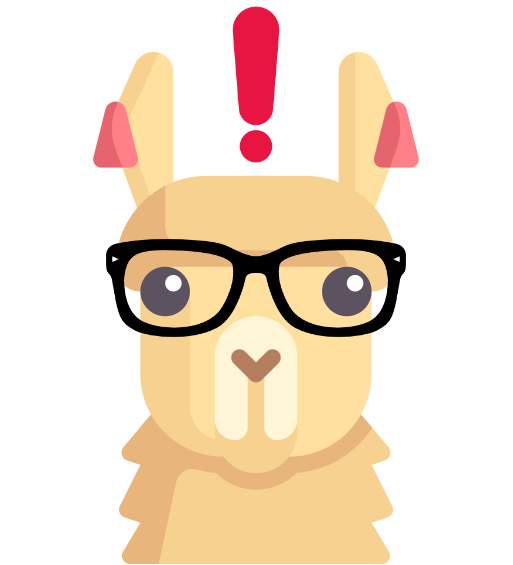}AHA: A Vision-Language-Model for Detecting and Reasoning Over Failures in Robotic Manipulation}

\author{ Jiafei Duan~$^{1,2}$ 
\hspace{4px} {Wilbert Pumacay}$^{3}$ \hspace{4px} {Nishanth Kumar}$^{1,4}$ \hspace{4px} Yi Ru Wang~$^{1,2}$ \\
\textbf{Shulin Tian~$^5$} \hspace{4px} \textbf{Wentao Yuan~$^{1,2}$} \hspace{4px} \textbf{Ranjay Krishna~$^{2, 6}$} \hspace{4px} \textbf{Dieter Fox~$^{1, 2}$}
\\
{\textbf{Ajay Mandlekar}}$^{*1}$\hspace{4px} {\textbf{Yijie Guo}}\thanks{Equal advising} $^{\;1}$\\
$^1$NVIDIA \hspace{6px} 
$^2$University of Washington \hspace{6px}
$^3$Universidad Católica San Pablo \hspace{6px}\\
$^4$MIT \hspace{6px} 
$^5$Nanyang Technological University $^6$Allen Institute for Artificial Intelligence\hspace{6px}\\
[1em]
\large\textbf{\href{https://aha-vlm.github.io/}{aha-vlm.github.io}
}
}

\begin{document}
\maketitle

\begin{abstract}
Robotic manipulation in open-world settings requires not only task execution but also the ability to detect and learn from failures. While recent advances in vision-language models (VLMs) and large language models (LLMs) have improved robots' spatial reasoning and problem-solving abilities, they still struggle with failure recognition, limiting their real-world applicability. We introduce \name, an open-source VLM designed to detect and reason about failures in robotic manipulation using natural language. By framing failure detection as a free-form reasoning task, \name identifies failures and provides detailed, adaptable explanations across different robots, tasks, and environments. We fine-tuned \name using \texttt{FailGen}, a scalable framework that generates the first large-scale dataset of robotic failure trajectories, the \name dataset. \texttt{FailGen} achieves this by procedurally perturbing successful demonstrations from simulation. Despite being trained solely on the \name dataset, \name generalizes effectively to real-world failure datasets, robotic systems, and unseen tasks. It surpasses the second-best model (GPT-4o in-context learning) by 10.3\% and exceeds the average performance of six compared models—including five state-of-the-art VLMs—by 35.3\% across multiple metrics and datasets. We integrate \name into three manipulation frameworks that utilize LLMs/VLMs for reinforcement learning, task and motion planning, and zero-shot trajectory generation. \name’s failure feedback enhances these policies' performances by refining dense reward functions, optimizing task planning, and improving sub-task verification, boosting task success rates by an average of 21.4\% across all three tasks compared to GPT-4 models. 
\end{abstract}
\vspace{-1em}


\section{Introduction}
\label{intro}

In recent years, foundation models have made remarkable progress across various domains, demonstrating their ability to handle open-world tasks\cite{driess2023palm, alayrac2022flamingo, achiam2023gpt, zhang2023adding}. These models, including large language models (LLMs) and vision-language models (VLMs), have shown proficiency in interpreting and executing human language instructions\cite{ouyang2022training}, producing accurate predictions and achieving strong task performance. However, despite these advancements, key challenges remain—particularly with hallucinations, where models generate responses that deviate from truth. Unlike humans, who can intuitively detect and adjust for such errors, these models often lack the mechanisms for recognizing their own mistakes\cite{lin2021truthfulqa, chen2021evaluating, heyman2008children}.

Learning from failure is a fundamental aspect of human intelligence. Whether it’s a child learning to skate or perfecting a swing, the ability reason over failures is essential for improvement\cite{young2009learning, gopnik2020childhood, heyman2008children}. The concept of improvement through failures is widely applied in training foundation models and is exemplified by techniques such as Reinforcement Learning with Human Feedback (RLHF)\cite{ouyang2022training, christiano2017deep}, where human oversight and feedback steers models toward desired outcomes. This feedback loop plays a critical role in aligning generative models with real-world objectives. However, a crucial question persists: How can we enable these models to autonomously detect and reason about their own failures, particularly in robotics, where interactions and environments are stochastic and unpredictable?

This need is particularly pressing in robotics, where foundation models such as VLMs and LLMs are increasingly used to address open-world tasks. Recent advancements have enabled these models to tackle spatial reasoning, object recognition, and multimodal problem-solving—skills vital for robotic manipulation\cite{reid2024gemini, openai2024gpt4o, yuan2024robopoint, chen2024spatialvlm, wang2023newton}. VLMs and LLMs are already being integrated to automate reward generation for reinforcement learning\cite{ma2023eureka, ma2024dreureka}, develop task plans for motion planning\cite{curtis2024trustproc3ssolvinglonghorizon}, and even generate zero-shot robot trajectories\cite{huang2023voxposer, huang2024copa, duan2024manipulate, huang2024rekep}. While these models excel at task execution, they often face challenges in detecting and reasoning over failures—skills that are crucial for navigating dynamic and complex environments. For example, if a robot drops an object mid-task, a human observer would immediately recognize the error and take corrective action. How can we empower robots with similar capabilities, allowing them not only to perform tasks but also to detect and learn from their mistakes?

To learn from their mistakes, robots must first detect and understand why they failed. We introduce \name, an open-source vision-language model (VLM) that uses natural language to detect and reason about failures in robotic manipulation. Unlike prior work that treats failure reasoning as a binary detection problem, we frame it as a free-form reasoning task, offering deeper insights into failure mode reasoning. Our model not only identifies failures but also generates detailed explanations. This approach enables \name to adapt to various robots, camera viewpoints, tasks, and environments in both simulated and real-world scenarios. It can also be integrated into downstream robotic applications leveraging VLMs and LLMs. We make the following three major contributions:


\textbf{1. We introduce \texttt{FailGen}, a data generation pipeline for the procedural generation of failure demonstration data for robotic manipulation tasks across simulators.} 
To instruction-tune \name, we developed \texttt{FailGen}, the first automated data generation pipeline that procedurally creates the \name dataset—a large-scale collection of robotic manipulation failures with over 49K+ image-query pairs across 79 diverse simulated tasks. Despite being fine-tuned only on the \name dataset, \name demonstrates strong generalization to real-world failure datasets, different robotic systems, and unseen tasks, as evaluated on three separate datasets not included in the fine-tuning. \texttt{FailGen} is also flexible data generation pipeline integrates seamlessly with various simulators, enabling scalable procedural generation of failure demonstrations.

\textbf{2. We demonstrate that \name excels in failure reasoning, generalizing across different embodiments, unseen environments, and novel tasks, outperforming both open-source and proprietary VLMs.} 
Upon fine-tuning \name, we benchmarked it against six state-of-the-art VLMs, both open-source and proprietary, evaluating performance across four metrics on three diverse evaluation datasets, each featuring different embodiments, tasks, and environments out-of-distribution from the training data. \name outperformed GPT-4o model by more than 20.0\% on average across datasets and metrics, and by over 43.0\% compared to LLaVA-v1.5-13B \cite{liu2023improvedllava}, the base model from which \name is derived. This demonstrates \name's exceptional ability to detect and reason about failures in robotic manipulation across embodiment and domains.


\textbf{3. We show that \name enhances downstream robotic applications by providing failure reasoning feedback.} We demonstrate that \name can be seamlessly integrated into robotic applications that utilize VLMs and LLMs. By providing failure feedback, \name improves reward functions through Eureka reflection, enhances task and motion planning, and verifies sub-task success in zero-shot robotic manipulation. Across three downstream tasks, our approach achieved an average success rate 21.4\% higher than GPT-4 models, highlighting \name's effectiveness in delivering accurate natural language failure feedback to improve task performance through error correction.

\begin{figure}[t]
    \centering
    \includegraphics[width=\textwidth]{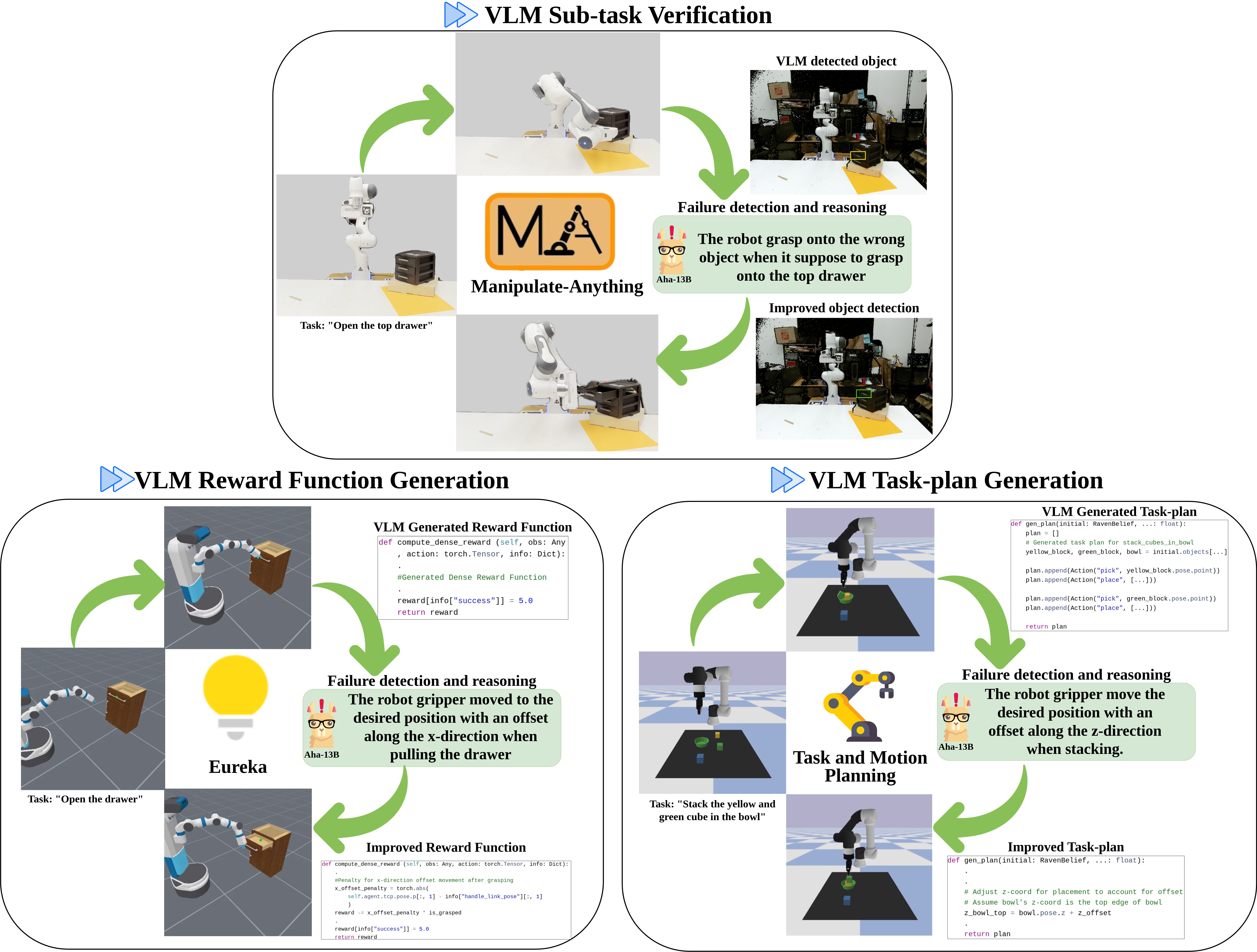}
    \caption{\name is a Vision-Language Model designed to detect and reason about failures in robotic manipulation. As an instruction-tuned VLM, it can enhance task performance in robotic applications that utilize VLMs for reward generation, task planning, or sub-task verification. By incorporating \name into the reasoning pipeline, these applications can achieve accelerated and improved performance.}
    \label{teaser}
\end{figure}

\section{Related Work}
\label{related}

\name enables language reasoning for failure detection in robotic manipulation, enhancing downstream robotics applications. To provide context, we review progress in: 1) failure detection in robotic manipulation, 2) data generation in robotics, and 3) foundation models for robotic manipulation.

\textbf{Failure Detection in Robotic Manipulation}. Failure detection and reasoning have long been studied in the Human-Robot Interaction (HRI) community \cite{ye2019human, khanna2023user} and in works leveraging Task and Motion Planning (TAMP) \cite{garrett2020pddlstream}. With the recent widespread adoption of LLMs and VLMs in robot manipulation systems—either for generating reward functions or synthesizing robot trajectories \cite{ma2023eureka, ma2024dreureka} in a zero-shot manner—the importance of detecting task failures has regained prominence \cite{huang2023voxposer,duan2024manipulate,skreta2024replan,ha2023scaling}. Most modern approaches focus on using off-the-shelf VLMs or LLMs as success detectors \cite{ma2022vip,ha2023scaling,wang2023gensim,duan2024manipulate}, and some employ instruction-tuning of VLMs to detect failures \cite{du2023vision}. However, these methods are often limited to binary success detection and does not provide language explanations for why failures occur. Our framework introduces failure reasoning in a new formulation, generating language-based explanations of failures to aid robotics systems that leverage VLMs and LLMs in downstream tasks.

\textbf{Data Generation in Robotics}
There have been many methods in robotic manipulation that automate data generation of task demonstrations at scale \cite{mandlekar2023mimicgen,hoque2024intervengen}, whether for training behavior cloning policies, instruction-tuning VLMs \cite{yuan2024robopoint}, or curating benchmarks for evaluating robotic policies in simulation \cite{xie2024decomposing,pumacay2024colosseum}. A well-known example is MimicGen \cite{mandlekar2023mimicgen}, which automates task demonstration generation via trajectory adaptation by leveraging known object poses. Additionally, works like RoboPoint use simulation to generate general-purpose representations for robotic applications, specifically for fine-tuning VLMs. Similarly, systems like The Colosseum \cite{pumacay2024colosseum} automate data generation for curating benchmarks in robotic manipulation. Our approach aligns closely with RoboPoint, as we also leverage simulation to generate data for instruction-tuning VLMs. However, unlike RoboPoint, we focus on synthesizing robotic actions in simulation rather than generating representations like bounding boxes or points.

\textbf{Foundation Models for Robotic Manipulation.}
In recent years, leveraging foundation models for robotic manipulation has gained significant attention due to the effectiveness of LLMs/VLMs in interpreting open-world semantics and their ability to generalize across tasks \cite{duan2022survey,hu2023toward, firoozi2023foundation,urain2024deep}. Two main approaches have emerged: the first uses VLMs and LLMs in a promptable manner, where visual prompts guide low-level action generation based on visual inputs \cite{liu2024moka,huang2024copa,huang2024rekep}. The second focuses on instruction-tuning VLMs for domain-specific tasks \cite{li2024llara}. For example, RoboPoint \cite{yuan2024robopoint} is tuned for spatial affordance prediction, and Octopi \cite{yu2024octopi} for physical reasoning using tactile images. These models generalize beyond their training data and integrate seamlessly into manipulation pipelines. Our approach follows this second path, developing a scalable method for generating instruction-tuning data in simulation and fine-tuning VLMs specialized in detecting and reasoning about robotic manipulation failures, with applications that extend beyond manipulation tasks to other robotic domains.


\section{The \name Dataset}
\label{data}



We leveraged \texttt{FailGen} to procedurally generate the \name dataset from RLBench tasks \citep{james2020rlbench} and used it for the instruction-tuning of \name. In this section, we begin by categorizing common failure modes in robotics manipulation and defining a taxonomy of failures in Section \ref{subsection:taxonomy}. Next, we explain how this taxonomy is used with \texttt{FailGen} to automate the data generation for the \name dataset in simulation in Section \ref{robofail}.

\begin{table}
    \centering
    \captionof{table}{\textbf{\name datasets for instruction-tuning.} We combined the \name dataset, our large-scale robotic manipulation failure dataset, with VQA and object detection data. By incorporating this diverse data mix into the fine-tuning process, \name is able to reason about failures in robotic manipulation across different domains, embodiments, and tasks.}
    \scriptsize
    \begin{tabularx}{\linewidth}{lXXX}
        \toprule
        Source & The \name dataset (Train) & VQA \cite{liu2023improvedllava} & LVIS \cite{gupta2019lvis} \\ \midrule
        & \includegraphics[width=\linewidth]{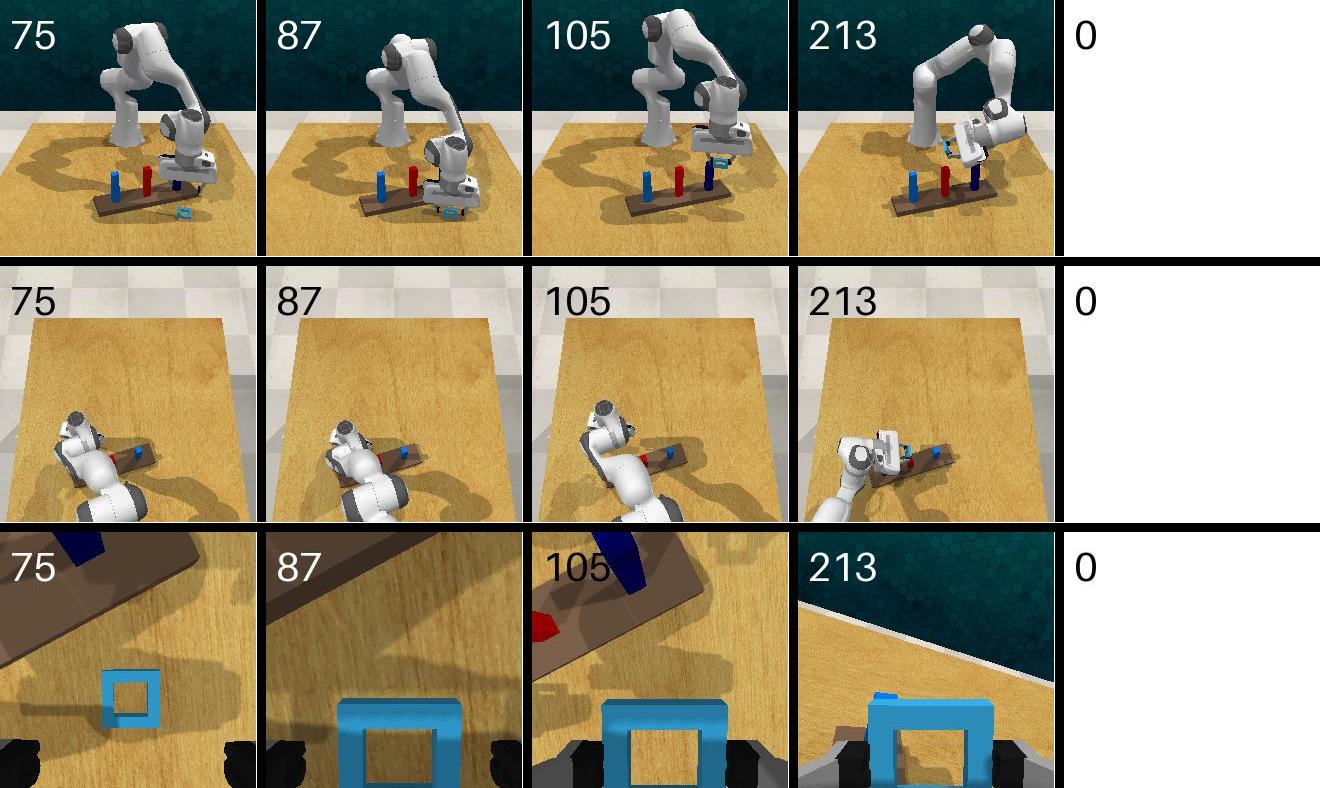} & \includegraphics[width=0.8\linewidth]{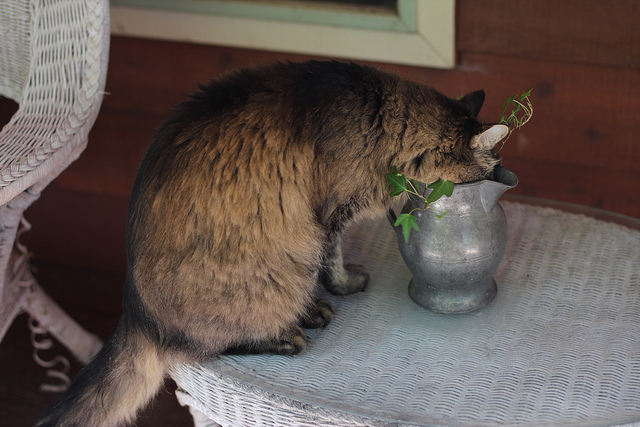} & \includegraphics[width=0.8\linewidth]{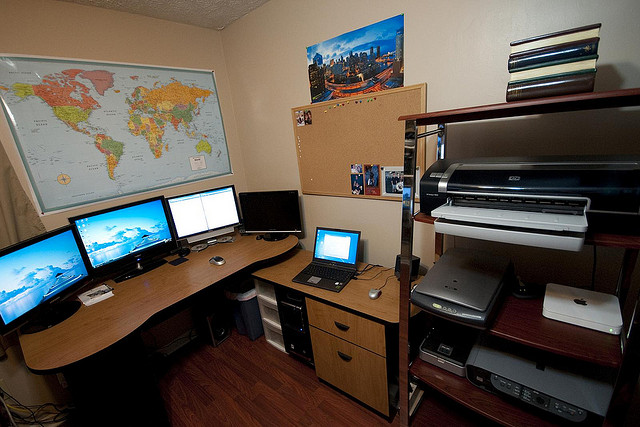} \\ \midrule
        Quantity & 49K & 665K & 100K \\ \midrule
        Query & For the given sub-tasks, first determine it has succeed by choosing from ["yes", "no"] and then explain the reason why the current sub-tasks has failed. &  What is the cat doing in the image? & 
        Find all instances of drawer.\\ \midrule
        Answer & No, The robot gripper rotated with an incorrect roll angle & The cat is sticking its head into a vase or container, possibly drinking water or investigating the interior of the item. &  [(0.41, 0.68, 0.03, 0.05), (0.42, 0.73, 0.04, 0.08), ...]  \\ \bottomrule
    \end{tabularx}
    \label{tab:data_mix}
\end{table}
\subsection{Failure Modes in Robotic Manipulation}
\label{subsection:taxonomy}

To curate an instruction-tuning dataset of failure trajectories for robotic manipulation tasks, we began by systematically identifying prevalent failure modes. Our approach involved a review of existing datasets, including DROID \cite{khazatsky2024droid} and Open-X Embodiment \cite{padalkar2023open}, as well as an analysis of policy rollouts from behavior cloning models. We examined failures occurring in both teleoperated and autonomous policies. Building upon prior works, such as REFLECT \cite{liu2023reflect}, we formalized a taxonomy encompassing seven distinct failure modes commonly observed in robotic manipulation: incomplete grasp, inadequate grip retention, misaligned keyframe, incorrect rotation, missing rotation, wrong action sequence, and wrong target object.

\textbf{Incomplete Grasp (\texttt{No\_Grasp}) Failure:} \texttt{No\_Grasp} is an object-centric failure that occurs when the gripper reaches the desired grasp pose but fails to close before proceeding to the next keyframe.

\textbf{Inadequate Grip Retention (\texttt{Slip}) Failure:} \texttt{Slip} is an object-centric failure that happens after the object has been successfully grasped. As the gripper moves the object to the next task-specific keyframe, the grip loosens, causing the object to slip from the gripper.

\textbf{Misaligned keyframe (\texttt{Translation}) Failure:} This action-centric failure occurs when the gripper moves toward a task keyframe, but a translation offset along the X, Y, or Z axis causes the task to fail.

\textbf{Incorrect Rotation (\texttt{Rotation}) Failure:} \texttt{Rotation} is an action-centric failure that occurs when the gripper reaches the desired translation pose for the sub-task keyframe, but there is an offset in roll, yaw, or pitch, leading to task failure.

\textbf{Missing Rotation (\texttt{No\_Rotation}) Failure:} \texttt{No\_Rotation} is an action-centric failure that happens when the gripper reaches the desired translation pose but fails to achieve the necessary rotation (roll, yaw, or pitch) for the sub-task, resulting in task failure.

\textbf{Wrong Action Sequence (\texttt{Wrong\_action}) Failure:} \texttt{Wrong\_action} is an action-centric failure that occurs when the robot executes actions out of order, performing an action keyframe before the correct one. For example, in the task \texttt{put\_cube\_in\_drawer}, the robot moves the cube toward the drawer before opening it, leading to task failure.

\textbf{Wrong Target Object (\texttt{Wrong\_object}) Failure:} \texttt{Wrong\_object} is an object-centric failure that occurs when the robot acts on the wrong target object, not matching the language instruction. For example, in the task \texttt{pick\_the\_red\_cup}, the gripper picks up the green cup, causing failure.

\begin{figure}[t]
    \centering
    \includegraphics[width=\textwidth]{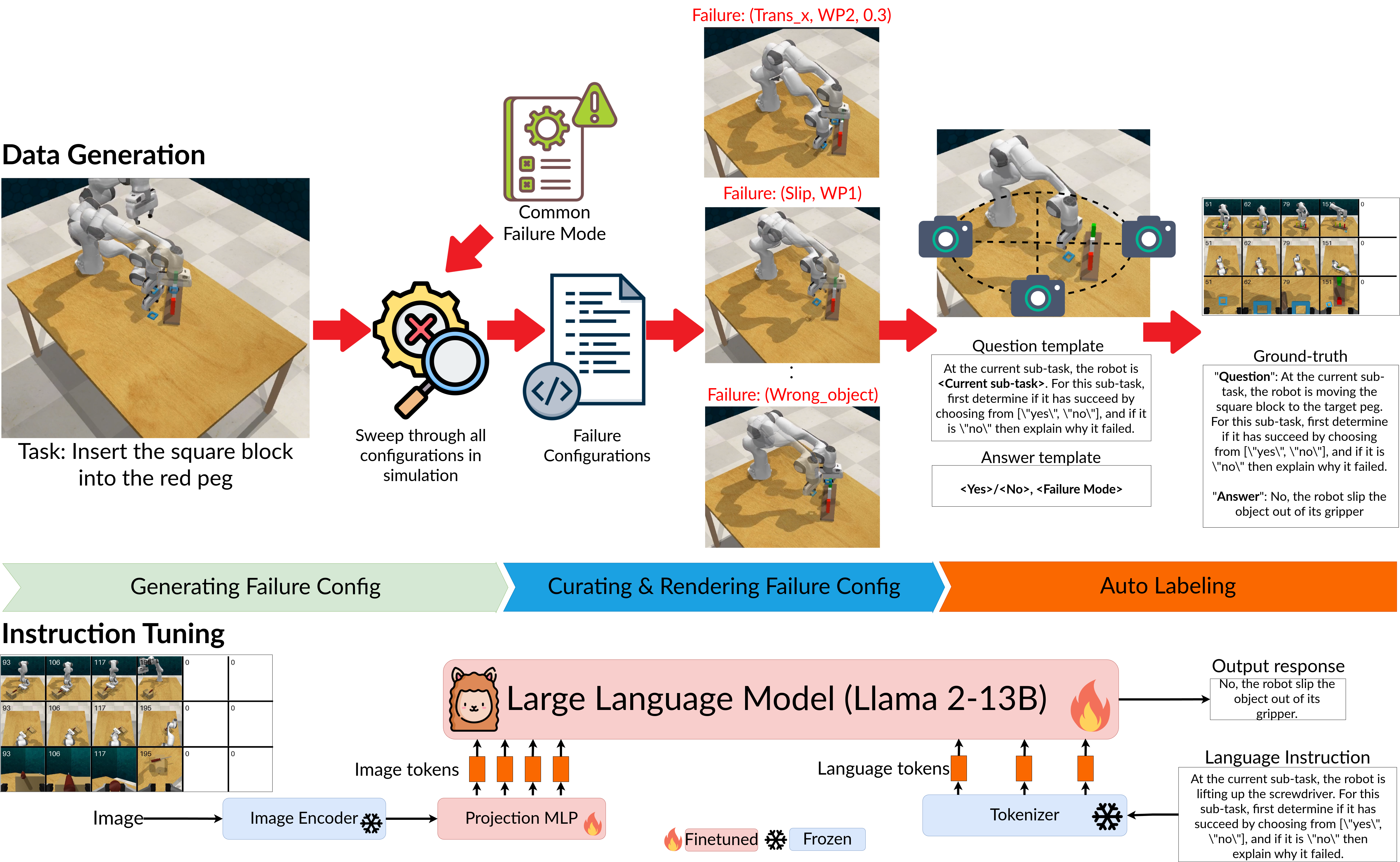}
    \caption{\textbf{Overview of \name Pipeline}. (Top) The data generation for \name is accomplished by taking a normal task trajectory in simulation and procedurally perturbing all keyframes using our taxonomy of failure modes. Through \texttt{FailGen}, we systematically alter keyframes to synthesize failure demonstrations conditioned on the original tasks. Simultaneously, we generate corresponding query and answer prompts for each task and failure mode, which are used for instruction-tuning. (Bottom) The instruction-tuning pipeline follows the same fine-tuning procedure as LLaVA-v1.5 \cite{liu2023improvedllava}, where we fine-tune only the LLM base model—in this case, LLaMA-2-13B and the projection linear layers, while freezing the image encoder and tokenizer.}
    \label{method}
\end{figure}

\subsection{Implementation of the \name dataset} 

\label{robofail}

The \name dataset is generated with RLBench \cite{james2020rlbench}, utilizing its keyframe-based formulation to dynamically induce failure modes during task execution. RLBench natively provides keyframes for task demonstrations, which enables flexibility in both object manipulation (handling tasks with varying objects) and the sequence of actions (altering the execution order of keyframes). Building on this foundation, we leverage \texttt{FailGen}, our custom environment wrapper to wrap around RLBench that allows for task-specific trajectory modifications through keyframes perturbations, object substitutions, and reordering of keyframe sequences. \texttt{FailGen} systematically generates failure trajectories aligned with the taxonomy defined in Section \ref{subsection:taxonomy}, yielding a curated dataset of 49k failure-question pairs.


To generate the \name dataset, we systematically sweep through all keyframes in each RLBench task, considering all potential configurations of the seven failure modes that could result in overall task failure. By leveraging the success condition checker in the simulation, we procedurally generate \texttt{YAML}-based configuration files by sweeping through each failure mode across all keyframes. These files provide details on potential failure modes, parameters (such as distance, task sequence, gripper retention strength, etc.), and corresponding keyframes that \texttt{FailGen} should perturb to induce failure. Additionally, we incorporate language templates to describe what the robot is doing between consecutive keyframes. Using these descriptions along with the failure modes, we can systematically curate question-answer pairs for each corresponding failure mode.


For specific failure modes, \texttt{No\_Grasp} is implemented by omitting gripper open/close commands at the relevant keyframes, effectively disabling gripper control. \texttt{Slip} introduces a timed release of the gripper shortly after activation. \texttt{Translation} and \texttt{Rotation} perturb the position and orientation of a keyframe, respectively, while \texttt{No\_Rotation} constrains the keyframe's rotational axis. \texttt{Wrong\_Action} reorders keyframe activations to simulate incorrect sequencing, and \texttt{Wrong\_Object} reassigns the keyframes intended for one object to another, maintaining the relative pose to mimic improper object manipulation. Using this pipeline, we also successfully generated a failure dataset from ManiSkill \citep{mu2021maniskill} and adapted RoboFail \citep{liu2023reflect} for the evaluation of \name. This further demonstrates the generalizability and versatility of \texttt{FailGen} in generating failure cases across different simulation environments.

\section{Method}
\label{sec:method}




This section outlines the failure reasoning problem formulation (Sec.\ref{sec3.1}) used to fine-tune and evaluate \name. Next, we discuss the curated data mix used for co-finetuning \name (Sec.\ref{sec3.2}). Finally, we detail the instruction fine-tuning pipeline and the model architecture selection for \name (Sec.\ref{sec3.3}).



\subsection{Failure Reasoning Formulation}
\label{sec3.1}

Unlike previous works \cite{liu2023reflect,skreta2024replan,duan2024manipulate} that primarily focus on detecting task success as binary classification problem, we approach failure reasoning by first predicting a binary success condition ("Yes" or "No") of the given sub-task based on a language specification and an input image prompt. If the answer is "No", the VLM is expected to generate a concise, free-form natural language explanation detailing why the task is perceived as a failure. To formulate failure reasoning, we prompt the VLMs to analyze the trajectory failures at the current sub-task and provide reasoning for \emph{why} or \emph{what} led to the failure. We define manipulation task trajectories as a series of sub-tasks $\{S_0, S_1, S_2, \dots, S_t\}$, where each sub-task is represented by two consecutive keyframes. For example, in a task like "stacking cubes", a sub-task could represent a primitive action, such as 'picking up the cube'.
For the input formulation in VLMs for instruction fine-tuning and evaluation, we required a query prompt with an input image for prompting the VLMs. The query prompt was generated using a template corresponding to the current sub-task the robot is performing. To capture the temporal relationships within the action sequence, the input image was constructed by selecting a single frame that represents the robot's trajectory up to the current sub-task and concatenating it with frames from other viewpoints in the rollout sequence, as shown in Table \ref{tab:data_mix}.

This input frame is built by concatenating all keyframes up to the current sub-task in temporal order, from left to right, with any remaining keyframes replaced by white image patches. To mitigate occlusions, we also included all the available camera viewpoints, concatenating them alongside the temporal sequence, and provide a detailed task description in the prompt, as illustrated in Table \ref{tab:data_mix} (left image). The image data is structured as a matrix $\mathbf{I}$, where each row corresponds to a different camera viewpoint $\{V_0, V_1, \dots, V_n\}$ and each column captures the temporal sequence of keyframes $\{S_0, S_1, S_2, \dots, S_t\}$.
This formulation for curating images serves as a general approach for formatting all datasets used for fine-tuning and evaluation. This structured input enables consistent handling of data across different tasks and viewpoints.
Overall, our failure reasoning problem is to prompt VLM with sub-task discription and keyframe trajectory image to predict the success condition and language description of failure reason for each sub-task, as shown in Table \ref{tab:data_mix}.

\subsection{Synthetic Data for Instruction-tuning}
\label{sec3.2}

To facilitate the instruction-tuning of \name, we needed to systematically generate failure demonstration data. To achieve this, we developed \texttt{FailGen}, an environment wrapper that can be easily applied to any robot manipulation simulator. \texttt{FailGen} systematically perturbs successful robot trajectories for manipulation tasks, transforming them into failure trajectories with various modes of failure as depicted in Figure \ref{method} (Top image). Using \texttt{FailGen}, we curated the \name dataset (Train) dataset by alternating across 79 different tasks in the RLBench simulator, resulting in 49k failure image-text pairs. Furthermore, following proper instruction-tuning protocols for VLMs \cite{liu2023improvedllava} and building on prior works \cite{brohan2023rt,yuan2024robopoint}, co-finetuning is crucial to the success of instruction fine-tuning of VLMs. Therefore, in addition to the \name dataset, we co-finetuned \name with general visual question-answering (VQA) datasets sourced from internet data, which helps models retain pre-trained knowledge. Specifically, we included the VQA dataset \cite{liu2023improvedllava}, containing 665k conversation pairs, and the LVIS dataset \cite{gupta2019lvis}, which comprises 100k instances with predicted bounding box centers and dimensions, as summarized in Table \ref{tab:data_mix}.

\subsection{Instruction Fine-tuning}
\label{sec3.3}

We followed the instruction-tuning pipeline outlined by \citep{liu2023llava}. As depicted in Fig.~\ref{method}, our model architecture includes an image encoder, a linear projector, a language tokenizer, and a transformer-based language model. The image encoder processes images into tokens, projected by a 2-layer linear projector into the same space as the language tokens. These multimodal tokens are then concatenated and passed through the language transformer. All components are initialized with pre-trained weights. During fine-tuning, only the projector and transformer weights are updated, while the vision encoder and tokenizer remain frozen. The model operates autoregressively, predicting response tokens and a special token marking the boundary between instruction and response. 

\section{Experimental Results}
\label{result}


In this section, we evaluate \name's detection and reasoning performance against six state-of-the-art VLMs, including both open-source and proprietary models, some utilizing in-context learning. The evaluation spans three diverse datasets, covering out-of-domain tasks, various simulation environments, and cross-embodiment scenarios. We then assess \name's ability to retain general world knowledge after fine-tuning on domain-specific data. Finally, we explore its potential to improve downstream robotic manipulation tasks.

\begin{table}[t]
\captionof{table}{\footnotesize{\textbf{Quantitative Evaluation on Failure Detection and Reasoning.} \method was evaluated and benchmarked against three open and three proprietary VLMs and one visual prompting baseline across three evaluation datasets. \method outperformed all other VLMs on every evaluation dataset and nearly every evaluation metric, with the exception of the \name (Test) dataset, where GPT-4o exceeded by less than 3\%.}}
    \centering
    \resizebox{\textwidth}{!}{%
    \begin{tabular}{@{}ccccccc@{}}
    \toprule
    \multirow{2}{*}{\textbf{Models}} & \multirow{2}{*}{\textbf{Evaluation Datasets}} & \multicolumn{4}{c}{\textbf{Evaluation Metrics}} \\
    \cmidrule(lr){3-6}
    & & $\text{ROUGE}_L$ $\uparrow$ & Cosine Similarity $\uparrow$ &  Binary Success(\%) $\uparrow$ & LLM Fuzzy Match $\uparrow$  \\
    \midrule
    \multirow{3}{*}{\textbf{LLaVA-v1.5-13B} \cite{liu2023improvedllava}}
        & \textbf{\name dataset (Test set)} & 0.061 & 0.208 & 0.080 & 0.648 \\
        & \textbf{ManiSkill-Fail} & 0.000 & 0.208 & 0.022 & 0.270\\
        &\textbf{RoboFail} \cite{liu2023reflect} & 0.000 & 0.203 & 0.000 & 0.404 \\
    \cmidrule(lr){2-6}
    \multirow{3}{*}{\textbf{LLaVA-NeXT-34B} \cite{liu2024llava}}
        & \textbf{\name dataset (Test set)} & 0.013 & 0.231 & 0.017 & 0.626 \\
        & \textbf{ManiSkill-Fail} & 0.001 & 0.195 & 0.007 & 0.277\\
        &\textbf{RoboFail} \cite{liu2023reflect} & 0.018 & 0.188 & 0.017 & 0.351 \\
    \cmidrule(lr){2-6}
    \multirow{3}{*}{\textbf{Qwen-VL} \cite{bai2023qwen}}
        & \textbf{\name dataset (Test set)} & 0.000 & 0.161 & 0.000 & 0.426 \\
        & \textbf{ManiSkill-Fail} & 0.037 & 0.301 & 0.116 & 0.034\\
        &\textbf{RoboFail} \cite{liu2023reflect}& 0.000 & 0.159 & 0.000 & 0.050 \\
    \midrule
    \multirow{3}{*}{\textbf{Gemini-1.5 Flash} \cite{reid2024gemini}}
        & \textbf{\name dataset (Test set)} & 0.120 & 0.231 & 0.371 & 0.566 \\
        & \textbf{ManiSkill-Fail} & 0.003 & 0.121 & 0.014 & 0.032\\
        &\textbf{RoboFail} \cite{liu2023reflect}& 0.000 & 0.042 & 0.000 & 0.393 \\
    \cmidrule(lr){2-6}
    \multirow{3}{*}{\textbf{GPT-4o}}
        & \textbf{\name dataset (Test set)} & 0.251 & 0.308 & 0.500 & \textbf{0.784} \\
        & \textbf{ManiSkill-Fail} & 0.142 & 0.335 & 0.688 & 0.453\\
        &\textbf{RoboFail} \cite{liu2023reflect}& 0.114 & 0.318 & 0.554 & 0.438 \\
    \cmidrule(lr){2-6}
    \multirow{3}{*}{\textbf{GPT-4o-ICL (5-shot)}}
        & \textbf{\name dataset (Test set)} & 0.226 & 0.380 & 0.611 & 0.776 \\
        & \textbf{ManiSkill-Fail} & 0.341 & 0.429 & 0.971 & 0.630\\
        &\textbf{RoboFail} \cite{liu2023reflect}& 0.236 & 0.429 & 0.571 & 0.418 \\
    \midrule
    \multirow{3}{*}{\textbf{AHA-7B}}
        & \textbf{\name dataset (Test set)} & 0.434 & 0.574 & 0.691 & 0.695 \\
        & \textbf{ManiSkill-Fail} & \textbf{0.609} & 0.680 & 1.000 & 0.532\\
        &\textbf{RoboFail} \cite{liu2023reflect}& 0.204 & 0.394 & 0.625 & 0.439 \\
    \cmidrule(lr){2-6}
    \multirow{3}{*}{\textbf{AHA-13B (Ours)}}
        & \textbf{\name dataset (Test set)} & \textbf{0.446} & \textbf{0.583} & \textbf{0.702} & 0.768 \\
        & \textbf{ManiSkill-Fail} & 0.600 & \textbf{0.681} & \textbf{1.000} & \textbf{0.633}\\
        &\textbf{RoboFail} \cite{liu2023reflect}& \textbf{0.280} & \textbf{0.471} & \textbf{0.643} & \textbf{0.465} \\
    \bottomrule
    \end{tabular}
    }
    \label{tab:main_result}
\end{table}

\begin{table}[ht]
\centering
\caption{\textbf{Quantitative Evaluation on Standard VQA Benchmarks.} \method performs on par with LLaVA-13B \cite{liu2023improvedllava}, the VLM from which \name adapts its fine-tuning strategy.}
\resizebox{\textwidth}{!}{
\begin{tabular}{lcccccccc}
\toprule
          & MMBench \cite{liu2023mmbench} & ScienceQA \cite{lu2022learn} & TextVQA \cite{singh2019towards} & POPE \cite{li2023evaluating} & VizWiz\cite{gurari2018vizwiz} \\ \midrule
        LLaVA-13B (LLama-2) \cite{liu2023improvedllava} & \textbf{67.70} & \textbf{73.21} & \textbf{67.40} & \textbf{88.00} & 53.01 \\
AHA-13B (LLama-2) & 65.20 & 71.94 & 65.20 & 85.74 & \textbf{53.45} \\ 
\bottomrule
\end{tabular}
}
\label{table:commonsense}
\end{table}

\subsection{Experimental Setup}

To quantitatively evaluate \name's\ detection and reasoning capabilities for failures in robotic manipulation, we curated two datasets and adapted an existing failure dataset for benchmarking. To ensure a fair comparison of free-form language reasoning, we also employed four different evaluation metrics to measure semantic similarity between sentences.

\textbf{Benchmarks.} We curated three datasets to evaluate \name's reasoning and failure detection capabilities, benchmarking against other state-of-the-art VLMs. The first dataset, \name dataset (Test), includes 11k image-question pairs from 10 RLBench tasks, generated similarly to the fine-tuning data via \texttt{FailGen} (Section \ref{robofail}) but without overlapping with the tasks from the finetuning dataset. It evaluates \name's ability to generalize to novel, out-of-domain tasks. The second dataset, ManiSkill-Fail, comprises 130 image-question pairs across four tasks in ManiSkill \cite{mu2021maniskill}, generated using \texttt{Failgen} wrapper on Maniskill simulator. This dataset assesses \name's performance in a different simulator and under changing viewpoints. Lastly, we adapted a failure benchmark from the RoboFail dataset \cite{liu2023reflect}, which features real-world robot failures in seven UR5 robot tasks. This allows for evaluation across simulation and real-world trajectories and across different embodiments. 

\begin{figure}[ht]
    \centering
    \begin{subfigure}[b]{0.40\linewidth}
        \includegraphics[width=\linewidth]{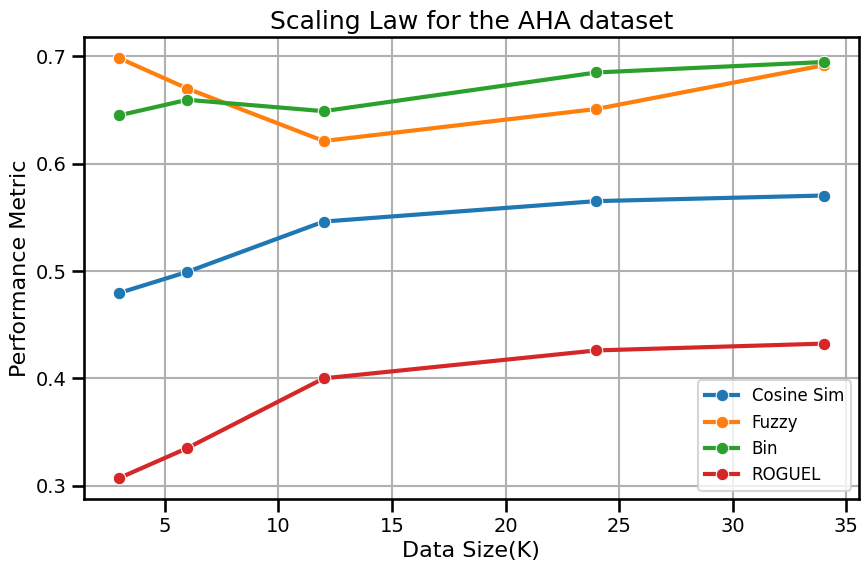}
    \end{subfigure}
    \begin{subfigure}[b]{0.55\linewidth}
        \includegraphics[width=\linewidth]{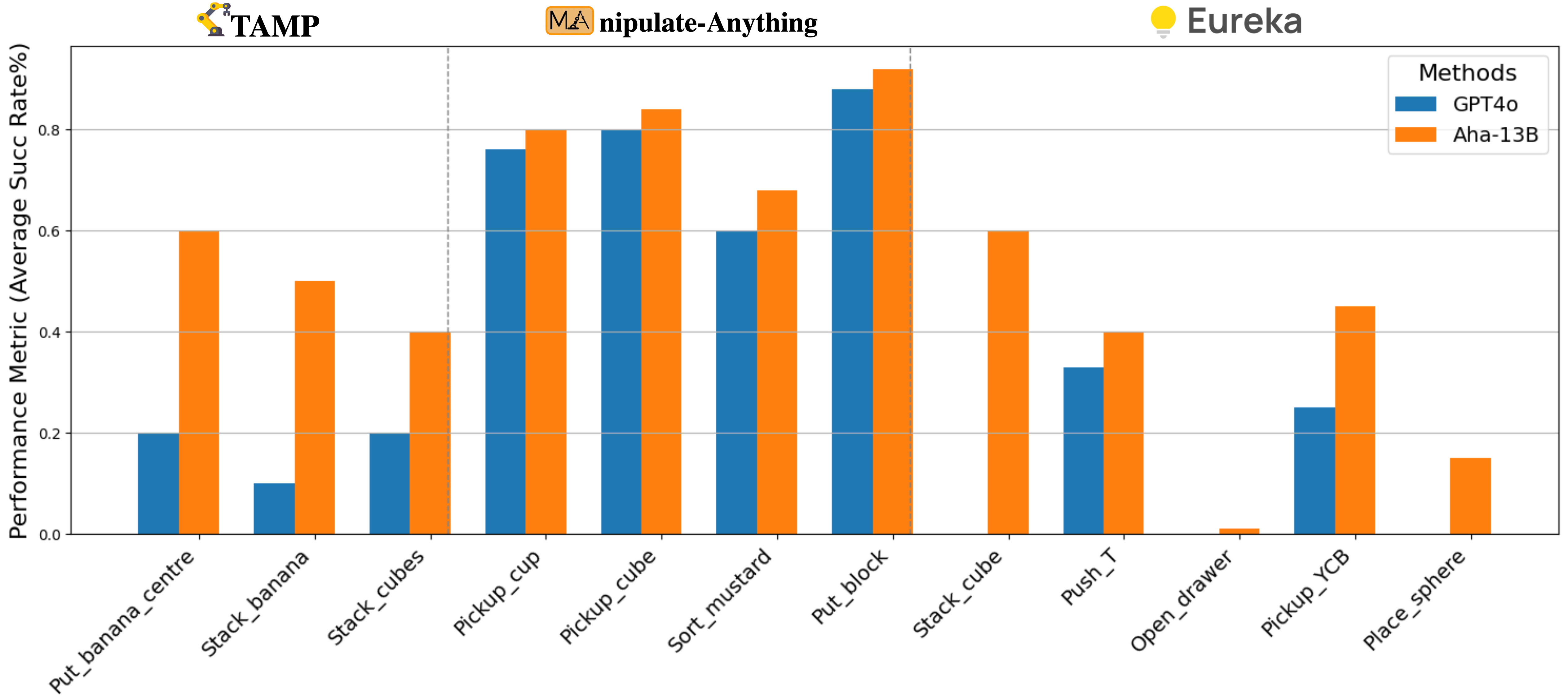}
    \end{subfigure}
    \caption{(Left) \textbf{Scaling law with the \name dataset}. Scaling of effect of model performance with varying domain specific fine-tuning data. (Right) \textbf{Downstream Robotic Application Performance.} \method outperforms GPT-4o in reasoning about failures within these robotic applications, leading to improved performance of the downstream tasks.}
    \label{fig:result_stuuff}
\end{figure}


\textbf{Evaluation Metrics.} To fairly evaluate success detection and free language reasoning across all datasets and baselines, we employ four metrics. First, the \textbf{ROUGE-L score} measures the quality of generated text by focusing on the longest common subsequence between candidate and reference texts. Second, we use \textbf{Cosine Similarity} to assess similarity between texts or embeddings, avoiding the "curse of dimensionality". Third, \textbf{LLM Fuzzy Matching} utilizes an external language model—specifically, Anthropic's unseen model, \texttt{claude-3-sonnet}—to evaluate semantic similarity in a teacher-student prompting format. Lastly, we calculate a \textbf{Binary success rate} by comparing the model's predictions directly against the ground truth for success detection.

\subsection{Quantitative Experimental Results}
We contextualize the performance of \name by conducting a systematic evaluation of failure reasoning and detection across these three datasets, general VQA datasets, and performed ablation studies.

\textbf{\name generalizes across embodiments, unseen environments, and novel tasks.} To ensure fairness and eliminate bias in the detection and reasoning capabilities of \name, we evaluated it on three different datasets that were never seen during fine-tuning, each designed to test a specific form of generalization. First, on the \name dataset (test) dataset, \name demonstrated its ability to \textbf{generalize reasoning across tasks and new behaviors within the same domain, outperforming the second-best performing VLM, GPT-4o ICL}, by an average margin of 12.6\% difference across all evaluation metrics. Second, we assessed \method on a dataset generated by the \texttt{Failgen} wrapper in a \textbf{different simulation domain}, ManiSkill, showing that our model outperforms GPT-4o-ICL by an average of 13.4\% difference across all metrics as depicted in Table \ref{tab:main_result}. Lastly, to demonstrate \textbf{generalization to real-world robots and different embodiments}, we evaluated \method on RoboFail~~\citep{liu2023reflect}, where it outperforms GPT-4o-ICL by 4.9\% difference.

\textbf{\name retains common sense knowledge.} We evaluated \method's performance on various VQA benchmarks and present the results in Table \ref{table:commonsense}
. \method \textbf{performs comparably to LLaVA-v1.5-13B (LLama-2) \cite{liu2023improvedllava} }, with only a 1.5\% margin difference as depicted in Table \ref{table:commonsense}. Notably, LLaVA-v1.5-13B is a VLM trained on the same pre-trained weights as \method but fine-tuned on VQA data. This indicates that \method is capable of functioning as a general purpose VLM, in addition to excelling at failure reasoning.


\textbf{\name's performance scales with data size.}
We evaluated Aha's performance using a range of \name data for instruction fine-tuning, spanning [3k, 6k, 12k, 34k, 48k, 60k], and co-trained individual checkpoints corresponding to these data sizes as shown in Figure \ref{fig:result_stuuff} (Left). The model was then assessed on the ManiSkill-Fail dataset across four evaluation metrics. An average quadratic fit gradient of 0.0022 across all four metrics demonstrates a \textbf{scaling effect with fine-tuning on our procedurally generated data pipeline}. This suggests that further scaling of the generated data may lead to improved model performance.

\setlength{\tabcolsep}{6pt}
\renewcommand{\arraystretch}{1}

\subsection{Downstream Robotics Tasks}
\label{tasks}

\begin{figure}[t]
    \centering
    \includegraphics[width=\textwidth]{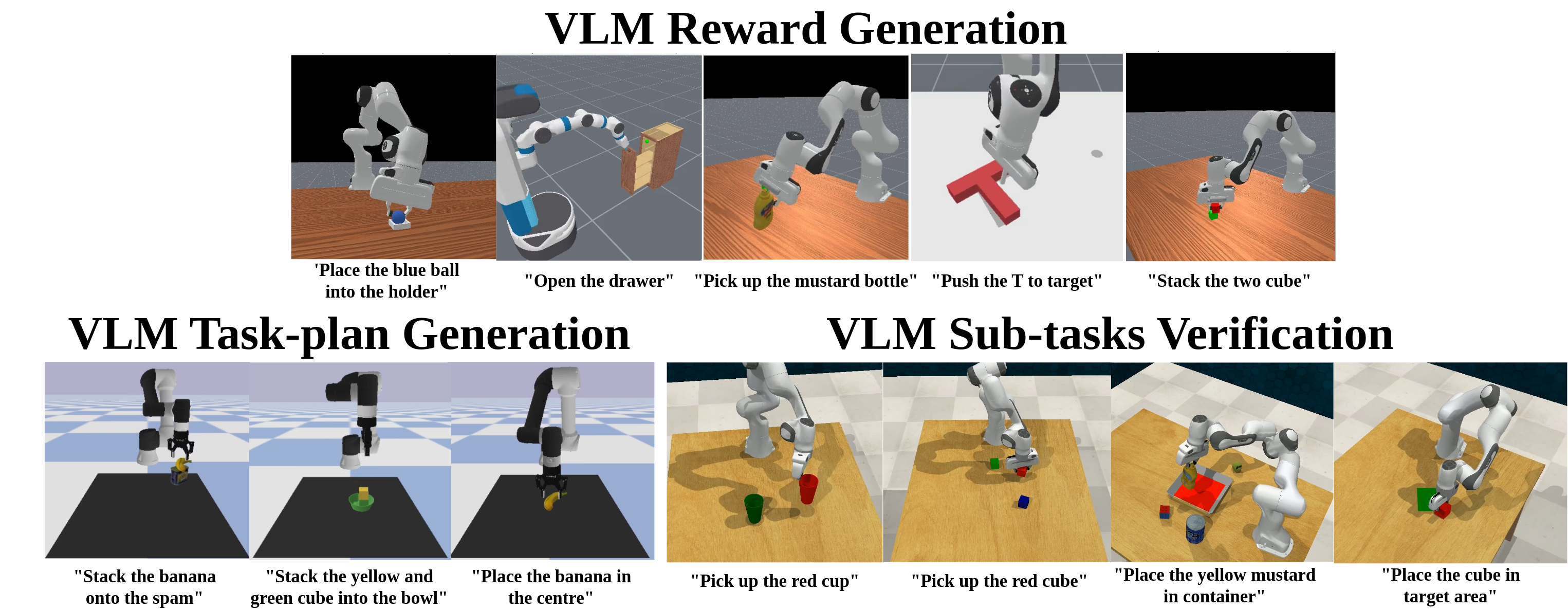}
    \caption{\textbf{Downstream Robotic Application.} We demonstrated that \name can be integrated into existing LLM/VLM-assisted robotic applications to provide failure reasoning and feedback, helping to accelerate and improve task success rates in these systems. }
    \label{task}
\end{figure}


We demonstrate that \name's\ failure detection and reasoning capabilities are useful across a wide spectrum of downstream robotics applications. This includes automatic reward generation for reinforcement learning applications \cite{ma2023eureka}, automatic task plan generation for task and motion planning applications \cite{curtis2024trustproc3ssolvinglonghorizon}, and as an improved verification step for automatic data generation systems \cite{duan2024manipulate}. Videos and detailed improved reward function, task plan, example videos from each applications and etc can be found on the project page: \href{https://aha-vlm.github.io/}{\textcolor{blue}{aha-vlm.github.io}}.

\textbf{\name enables efficient reward synthesis for reinforcement learning.} To evaluate this downstream task, we adapted Eureka's \cite{ma2023eureka} implementation to the ManiSkill simulator, which offers more state-based manipulation tasks. We strictly followed the Eureka reward function generation and reflection pipeline, modifying it by incorporating perception failure feedback via either \method or GPT-4o (acting as a baseline) to enhance the original LLM reflection mechanism. Instead of only including a textual summary of reward quality based on policy training statistics for automated reward editing, we further incorporated explanations of policy failures based on evaluation rollouts. We evaluated our approach on five reinforcement learning tasks from ManiSkill, ranging from tabletop to mobile manipulation. To systematically assess the reasoning capabilities of different VLMs under budget constraints, we sampled one reward function initially and allowed for iterations over two sessions of GPT API calls. Each policy was trained using PPO over task-specific training steps and evaluated across 1,000 test steps. During policy rollouts, we employed either \method or GPT-4o for reward reflection to improve the reward function. Comparing the evaluated policy success rates using different failure feedback VLMs, we observed that \method provided intuitive, human-level failure reasoning 
that aided in modifying and improving generated dense reward functions. This resulted in success across all five tasks within the budget constraints, and our approach \textbf{outperformed GPT4o by a significant margin of 22.34\% in task success rate} shown in Figure \ref{fig:result_stuuff} (Right).


\textbf{\name refines task-plan generation for TAMP.}
To demonstrate \name's\ utility within a planning system, we incorporated our approach into PRoC3S~\cite{curtis2024trustproc3ssolvinglonghorizon}.
The PRoC3S system solves tasks specified in natural language by prompting an LLM for a Language-Model Program (LMP) that generates plans, and then testing a large number of these plans within a simulator before executing valid plans on a robot.
If no valid plan can be found within a certain number of samples (100 in our experiments), the LLM is re-prompted for a new LMP given failure information provided by the environment.
Importantly, as is typical of TAMP methods, the original approach checks for a finite set of failures (inverse kinematics, collisions, etc.) from the environment, and returns any sampled plan that does not fail in any of these ways.
We incorporated a VLM into this pipeline in two ways: (1) we prompt the VLM with visualizations of failed plan executions within the simulator, ask it to return an explanation for the failure, and feed this back to PRoC3S' LLM during the LMP feedback stage, (2) after PRoC3S returns a valid plan, we provide a visualization of this to the VLM and ask it to return whether this plan truly achieves the natural language goal, with replanning triggered if not.
We compared GPT-4o and \method as the VLM-based failure reasoning modules within this implementation of PRoC3S across three tasks (shown in Figure~\ref{task}). Each task was evaluated over 10 trials, with a maximum of 100 sampling steps and three feedback cycles provided by either GPT-4o or \method. The success rate for each task was recorded. As shown in Figure Figure \ref{fig:result_stuuff} (Right), utilizing \method for \textbf{failure reasoning significantly improved the task success rate and outperforming GPT-4o by a substantial margin of 36.7\%}.

\textbf{\name improves task verification for zero-shot robot data generation.} To demonstrate \name's utility in zero-shot robot demonstration generation, we integrated our approach into the \texttt{Manipulate-Anything} framework. This open-ended system employs various Vision-Language Models (VLMs) to generate diverse robot trajectories and perform a wide range of manipulation tasks without being constrained by predefined actions or scenarios. A critical component of \texttt{Manipulate-Anything} is its sub-task verification module, which analyzes past and current frames to decide whether a sub-task has been achieved before proceeding or re-iterating over the previous sub-task. We replaced the original VLM (GPT-4V) in the sub-task verification module with \method and evaluated performance across four RLBench tasks (Figure \ref{task}), conducting 25 episodes for each task. Our results show that \textbf{substituting the sub-task verification module's VLM with \name improved reasoning accuracy and overall task success by an average of 5\%}.


\section{Conclusion}
\textbf{Limitations.}
\name currently outputs language reasoning that is closely aligned with the failure scenarios in the fine-tuning data. However, there is an opportunity to output more open-ended failures, to cover those arising from modes outside of the ones included in the failure taxonomy.
Additionally, while \texttt{FailGen} systematically curates failure data from simulations, distilling large pretrained policies to perform diverse tasks in simulation and sampling failure modes would allow us to generate more open-ended failure examples, potentially enhancing the instruction-tuned performance of \name. 

\textbf{Conclusion.} \label{conclusion} We introduce \name, an open-source vision-language model that significantly enhances robots' ability to detect and reason about manipulation task failures using natural language. By framing failure detection as a free-form reasoning task, \name not only identifies failures but also provides detailed explanations adaptable to various robots, tasks, and environments. Leveraging \texttt{FailGen} and the curated \name dataset, we trained \name on a diverse set of robotic failure trajectories. Our evaluations show that \name outperforms existing models across multiple metrics and datasets. When integrated into manipulation frameworks, its natural language feedback greatly improves error recovery and policy performance compared to GPT-4 models. These results demonstrate \name's effectiveness in enhancing task performance through accurate error detection and correction.

\section{Acknowledgement}
Jiafei Duan is supported by the Agency for Science, Technology and Research (A*STAR) National Science Fellowship. Wilbert Pumacay is supported by grant 234-2015-FONDECYT from Cienciactiva of the National Council for Science, Technology and Technological Innovation (CONCYTEC-PERU). 

\bibliography{main}

\end{document}